\documentclass[runningheads,a4paper]{llncs}

\usepackage{multirow}
\usepackage[utf8]{inputenc}
\usepackage[english]{babel}
\usepackage{graphicx}
\usepackage{amsmath}
\usepackage{amssymb}
\usepackage{array}
\newcolumntype{P}[1]{>{\centering\arraybackslash}p{#1}}

\usepackage{url}
\urldef{\mailsa}\path|{alfred.hofmann, ursula.barth, %ingrid.haas, frank.holzwarth,|
\urldef{\mailsb}\path|anna.kramer, leonie.kunz, %christine.reiss, nicole.sator,|
\urldef{\mailsc}\path|erika.siebert-cole, %peter.strasser, lncs}@springer.com|   

\usepackage{color}

\newcommand{\comment}[1]  {  }

\begin{document}

\mainmatter  % start of an individual contribution
\title{Every Untrue Label is Untrue in its Own Way:
Controlling Error Type with the Log Bilinear Loss}
\titlerunning{Every Untrue Label is Untrue in its Own Way}

\author{Yehezkel S. Resheff\inst{1,2} 
\and Amit Mandelbaum\inst{1}
\and Daphna Weinshall\inst{1}}

\institute{School of Computer Science and Engineering, The Hebrew University of Jerusalem \\
\{heziresheff, amit.mandelbaum, daphna\}@cs.huji.ac.il
\and
Edmond and Lily Safra Center for Brain Sciences, The Hebrew University of Jerusalem }

\maketitle

\begin{abstract} 
Deep learning has become the method of choice in many application domains of machine learning in recent years, especially for multi-class classification tasks. The most common loss function used in this context is the cross-entropy loss, which reduces to the log loss in the typical case when there is a single correct response label. While this loss is insensitive to the identity of the assigned class in the case of misclassification, in practice it is often the case that some errors may be more detrimental than others. Here we present the bilinear-loss (and related log-bilinear-loss) which differentially penalizes the different wrong assignments of the model. We thoroughly test this method using standard models and benchmark image datasets. As one application, we show the ability of this method to better contain error within the correct super-class, in the hierarchically labeled CIFAR100 dataset, without affecting the overall performance of the classifier.  
\end{abstract} 

\section{Introduction}

Multi-class classification may well be the most studied machine learning problem, from both the theoretical and the practical aspects. The problem is often phrased as an optimization problem, where we seek a classifier (or algorithm) which minimizes some loss function \cite{kotsiantis2007supervised}. To begin with we may phrase the problem using the 0-1 loss, which effectively counts the number of misclassified points. Most methods then replace this function by a surrogate loss due to various computational reasons (often just to achieve a tractable problem), or by the design of a more specific goal-oriented or application motivated loss function.

In this paper we go back to basics, and question the use of the 0-1 loss function in multi-class classification. Already in the binary scenario, we see the need for a more subtle loss function, which distinguishes between the 2 different types of possible errors: false negative - when a point of class 1 is wrongly classified as class 2, and false positive - when a point of class 2 is wrongly classified as class 1. When class 1 symbolizes the diagnosis of some serious illness while class 2 refers to the lack of findings, differential treatment of the two errors may be a matter of life and death. In most applications this distinction doesn't make it into the definition of the loss function (but see review of related work below), and it is usually approached post-hoc by such means as the ROC curve, which essentially describes the behavior of the classifier as a trade-off between the two types of error. 

These days the field of applied machine learning is swept by deep learning \cite{lecun2015deep}, and we customary solve very large multi-class classification problems with a growing number of classes. Now the question becomes more acute, and the 0-1 loss function does not seem to provide a good model for the real world anymore. In most application domains, different misclassification errors are likely to have different detrimental implications, and should be penalized accordingly to achieve an appropriate classifier. For example, the consequence of diagnosing a certain illness wrongly by coming up with a related illness may be considered less harmful to the patient than missing it altogether. 

Unlike binary classification, with multi-class problems we don't have only two types of errors to worry about, but rather $O(k^2)$ errors if there are $k$ labels to choose from. We can no longer postpone the resolution of the problem to some post-hoc stage, and the choice of a single threshold. Furthermore, when considering deep learning, encoding the desired trade-off between types of errors will hopefully lead to a representation more suitable to make the necessary distinctions. 

Yet in practice, the basic loss function one uses in the training of neural network classifiers has changed little, revealing its deep roots in the world of binary classification and the 0-1 loss function. Specifically, the categorical cross-entropy (which when using a single correct response reduces simply to log-loss), has the property of invariance to the division of the output weight among the erroneous labels, and therefore it is also invariant to the identity of the class an example is assigned to, in case of misclassification.

This is the problem addressed in this paper. We propose two loss functions, which provide the means for distinguishing between different misclassification errors by penalizing each error differently, while maintaining good performance in terms of overall accuracy.  Specifically, in Section~\ref{sec:loss}
\comment{
From its inception, the theory of machine learning has been predominantly occupied with a binary world. While some of the algorithmic and practical work is inherently multi-class (such as Decision Trees and related methods \cite{safavian1991survey}), a remarkable proportion of the most celebrated algorithmic developments take place in a binary setting.   

The most striking example of this is the overwhelming popularity of competition schemes for multi-class Support Vector Machine (SVM) classification, based on multiple binary classifiers, in spite of the existence of well founded, and inherently multi-class alternatives \cite{multi-crammer2002multisvm,multi-weston1998multi} (with more or less comparable performance  \cite{svm-compare,svm-compare-2}). 

Deep learning has swept the field of machine learning in recent years, and reached the status of the go-to method for many purposes, and especially for classification tasks. The common practice of using a final (soft-max) normalized output layer can again be viewed either as a truly multi-class method, or as a competition between many binary classifiers (in this view, each output unit is a binary classifier in the 1-versus-all sense, and the soft-max activation forces a competition between them). 

Either way, the element with deep roots in the binary setting is undoubtedly the loss function which is typically used. The categorical cross-entropy (which when using class labeling with a single correct response reduces simply to log-loss), has the property of invariance to the division of the output weight among the erroneous classes, and therefore it is also invariant to the identity of the class an example is assigned to, in the case of misclassification. 

While this property is often reasonable in the binary case (where there are only two types of errors), in the multi-class setting it may lead to very undesirable behavior. Namely, in real world problems, it is often the case that some errors are much worse than others, and we would thus prefer classifiers (all else being equal), that tend to make their errors between classes in a way that is controllable. 

In essence, this is an extension of the notion of type I and type II error, often used in binary decision problems, to the multi-class setting. Thus, for a problem with $k$ classes, we have $k\times(k-1)$ types of error. The method proposed here is then a means of trading-off these types of error, while maintaining the overall performance of the model. 

In this paper} we develop the bilinear loss (and related log-bilinear loss) which directly penalize a deep learning model differentially for weights assigned in the output layer, depending on both the correct label and the identity of the wrong labels which have weight assigned to them. 

In Section~\ref{section:experiments} we describe the empirical evaluation of training deep networks with these loss functions, using standard deep learning models and benchmark multi-class datasets. Thus we empirically show the ability of these methods to maintain good overall performance while redistributing the error differently between the possible wrong assignments. 
 
In Section~\ref{sec:hierar} we show an application of the proposed method for controlling error within the correct super-class in the hierarchically labeled CIFAR100 dataset, without significant reduction in the total accuracy of the final classifier (in fact, slight improvement is observed in our empirical evaluation). 

The contribution of this paper is twofold. First, we introduce the bilinear/log-bilinear loss functions and show their utility for controlling error location in deep learning models. Second, we suggest the task of confining error to the correct super-class in a hierarchical settings, and show how these loss functions can achieve this goal.   

\subsubsection*{Related Work}

Asymmetric loss functions have been studied extensively in the context of binary classification, both theoretically and algorithmically, see e.g.  \cite{bach2006considering,lin2002support,scott2012calibrated,masnadi2010risk}. Thus asymmetric loss functions have been shown to improve classification performance with asymmetrical error costs or imbalanced data. In a related approach, boosting methods (e.g. \cite{fan1999adacost}) are based on the asymmetric treatment of training points, and can naturally be modified to include cost sensitive loss functions \cite{sun2007cost}.  The optimal selection (with respect to a differential cost of the two types of error) of the final classifier's threshold has been discussed in \cite{karakoulas1998optimizing,koyejo2014consistent}, for example. This question also received specific attention in the context of the design of cascades of detectors \cite{wu2005linear,wu2008fast}, since as part of the construction of the cascade one must give the different errors - miss the event or detect it when it is absent - different weights in the different levels of the cascade.

In the context of multiclass classification, proper rescaling methods are discussed in \cite{zhou2010multi} in order to address the issue of imbalanced datasets; rescaling methods, however, typically penalize error based on the identify of the wrong label only. \cite{domingos1999metacost} describes a wrapper method to transform every multi-class classifier to a cost-sensitive one, while \cite{margineantu1999learning} shows how multi-class decision trees can be trained to approximate a general loss matrix. A truly asymmetric cost-sensitive loss function is used in \cite{lee2004multicategory}, employing a \emph{generalized cost matrix} somewhat similar to matrix $A$ defined in Section~\ref{sec:loss}, while \cite{zhang2010cost} offers a formulation which is based on the Bayes-decision theory and the $k$-nearest neighbor classifier.

In the context of deep learning, the purpose of a cost sensitive objective is not only to change the output of the model, but first and foremost to learn a representation in intermediate layers of the network that is able to better capture the aspects of the data that are important according to the relative cost of the different types of error. This is demonstrated in the hierarchical experiments below (Section \ref{section:experiments}).

\section{The [Log] Bilinear-loss}
\label{sec:loss}

We assume a model with $k$ non-negative output units, such that the output for the $i-th$ example is:

\[  \hat{y}^{(i)} = \hat{y}^{(i)}_1, \dots , \hat{y}^{(i)}_k \]

\noindent and further assume a per-example normalized output, i.e.: $ \forall i: \sum_{j=1}^{k} \hat{y}^{(i)}_j = 1 $

In most cases in practice, the training set is labeled with a single correct response per example. Thus, the common cross-entropy loss reduces to $-log(y^{(i)}_{l_i})$, where $l_i$ is the correct label for the $i$'th example. This loss essentially represents an implicit policy of rewarding for weight placed on the correct answer, while being indifferent to the identity of the wrong labels which have weight assigned to them.

Arguably, this common practice is often in direct opposition to the goal of the process of learning; in many real world scenarios, some mistakes are extremely costly while others are of little consequence. In this section we develop alternative loss functions which address this issue directly. %(these will then be used in conjunction with the regular log-loss and demonstrated in section (\ref{section:experiments}) below). 

We start by modifying the loss function used for training the classifier. We augment the usual loss function with a term where a non-negative cost is assigned to any wrong classification, and where the cost is based on \emph{both} the correct label and the identity of the misclassified label in an asymmetrical manner. Specifically, let $a_{i,j}$ denote the relative cost associated with assigning the label $j$ to an example whose correct label is $i$, and let $A=\{ a_{i,j}\}\in{\cal R}^{k\times k}$ denote the penalty matrix.

Using $A$ we define two related loss functions: The Bilinear loss is defined as:
\begin{equation}
	L_{B} =  y^T A \hat{y} 
    \label{eq:1}
\end{equation}
where $y$ denotes the correct output ($y$ is a probability vector). Similarly, the log-Bilinear loss is defined as:
\begin{equation}
\label{Log-Bilinear-Loss}
	L_{LB} =  -y^T A log( 1 - \hat{y})  
\end{equation}
where $log(\cdot)$ operates element-wise. 

Finally, we combine the regular cross-entropy loss with (\ref{eq:1}) and (\ref{Log-Bilinear-Loss}) to achieve a loss function with the known benefits of cross-entropy, and which also provides the deferential treatment of errors:
\begin{equation}
\label{eq:full-B}
	L_{CE+B} = (1-\alpha) L_{CE} + \alpha y^T A \hat{y} 
\end{equation}

\begin{equation}
\label{eq:full-LB}
	L_{CE+LB} = (1-\alpha) L_{CE} - \alpha y^T A log( 1 - \hat{y})  
\end{equation}
where $L_{CE} = -\sum y_ilog(\hat{y}_i)$ is the regular cross-entropy loss. %In the next section we evaluate these two formulations using benchmark datasets. 

The fundamental difference between the bilinear and log-bilinear loss formulations is the implied view on what in the output of the model is to be penalized (and thus controlled). The bilinear formulation adds a constant cost $a_{ij} \Delta p_j$ for each $\Delta p_j$ increase in the $j$'th normalized output unit, for a training example from the $i$'th class. The log-bilinear formulation on the other hand is insensitive to this sort of increase as long as the overall value in the $j$'th output is small, but picks up sharply as $p_j$ approaches $1$. 

Furthermore, for two equally penalized mistakes $j, k$ for an example from class $i$ (meaning $a_{ij} = a_{ik}$), the bilinear loss is insensitive to the division of error between the two classes, since $\alpha_1 a_{ij} + \alpha_2 a_{ik} = (\alpha_1+\alpha_2) a_{ij}$, so only the sum of the assignments to these two classes matters. The log-bilinear loss however amounts to $\alpha_1 log(1 - a_{ij}) + \alpha_2 log( 1- a_{ik})$, which for a constant sum is maximized when all the weight is placed on one of the errors. 

The meaning of this property is that the bilinear loss is penalizing for the total weight placed on erroneous classes weighted by the relative importance (as measured by the penalty matrix $A$). The log-bilinear loss is penalizing peakiness of the wrong assignment, again weighted by importance. Thus, the bilinear loss is a more likely candidate when we would like to control the locations where erroneous weights concentrate, and the log-bilinear loss when we would like to control which errors the model is confident about.  

We note that while we naturally apply the method to deep learning models in our experiments (section \ref{section:experiments}), these loss functions are applicable to any gradient based model with outputs as described above.

\section{Empirical evaluation of the loss functions}
\label{section:experiments}

In this section we evaluate and compare the efficacy of the two loss functions defined in (\ref{eq:full-B}) and (\ref{eq:full-LB}), in terms of two measures and the trade-off between them: (i) the total un-weighted error in the original classification task; (ii) the error distribution as measured by the different elements of the confusion matrix at train and test time.

Specifically, the purpose of the following experiments is: (a) to evaluate and compare the ability of the proposed formulation to control the location of error, using standard deep learning methods and benchmark datasets; (b) test the influence of the trade-off parameter $\alpha$ in the above formulations (\ref{eq:full-B}) and (\ref{eq:full-LB}); and (c) evaluate the trade-off between control of error location, and overall accuracy of the model. 

\subsection{Methods}

In order to test the ability to control the location of error, we randomly select a varying number $n$ of specific errors to be avoided. Each such error is defined by the identity of both the correct label and the wrong label (such as for example: "don't mistake a 2 for an 8"). In practice, this is done by sampling a random mask (Boolean matrix) of size $k^2$, with exactly $n$ positive off-diagonal elements, where $k$ is the number of classes. These $n$ locations are henceforth called \emph{the masked zone}. 

The datasets we use for empirical evaluation in this section are MNIST and CIFAR10, where $k=10$. For each combination of $n$ in $\{10, 20, 30, 40, 50\}$ and the trade-off parameter $\alpha$ in $\{0, .1, .5, .9, .95, .99\}$, we trained $50$ and $10$ models (MNIST and CIFAR10 respectively). Thus we trained a total of $1500$ MNIST models, and $300$ CIFAR10 models. This design was repeated for the Bilinear and Log-Bilinear loss functions, leading to a total of $3600$ models. 

All models were trained using the open source Keras and TensorFLow \cite{tensorflow2015-whitepaper} software packages \footnote{Materials available at: https://github.com/Hezi-Resheff/paper-log-bilinear-loss}. In order to facilitate the training of a large number of models, MNIST models were each trained for only $10$ epochs, and CIFAR10 models for $300$ epochs with an early stopping criterion. 

\subsection{MNIST dataset: results}

The MNIST dataset \cite{lecun1998gradient} is an old benchmark dataset of small images ($28\times28$ pixels) of hand-written digits. The data is divided into $55,000$ images in the training set, $5,000$ for validation, and $10,000$ images in the test set. In recent years, deep learning methods have been used successfully to reach almost perfect classification when using the MNIST dataset (see for example \cite{ciregan2012multi,jarrett2009best}).

\begin{table}[h!]
\centering
\caption{The model used in all MNIST experiments}
\label{tbl-model-MNIST}
\begin{tabular}{l|l|l}
  & Layer                 & \#params         \\ \hline \hline
  
1 & convolution (20; 5X5) & 520              \\
2 & max-pooling(2X2)      &                  \\
3 & dropout(20\%)         &                  \\ \hline
4 & convolution(50; 5X5)  & 25,050            \\
5 & max-pooling(2X2)      &                  \\
6 & dropout(20\%)         &                  \\ \hline
7 & fully-connected(500)  & 1,225,500          \\
8 & fully-connected(10)   & 5,010             \\ \hline \hline 
  & total:                & 1,256,080 
\end{tabular}
\end{table}

The model we use in our experiments (detailed in Table \ref{tbl-model-MNIST}) is a typical small model with two convolutional layers, followed by a single fully-connected layer. Max-pooling and dropout \cite{srivastava2014dropout} are used following each convolutional layer. This model was selected for the current experiments primarily because of the small training time required to achieve satisfactory results (approx. $99.2\%$ correct after $10$ epochs), which allowed us to generate a very large number of models, as discussed in the methods section above.  

\subsubsection*{Bilinear Loss.}

\begin{figure}[t!]
    \centering
    \includegraphics[trim={0 0 0 0},clip, width=1.\textwidth]{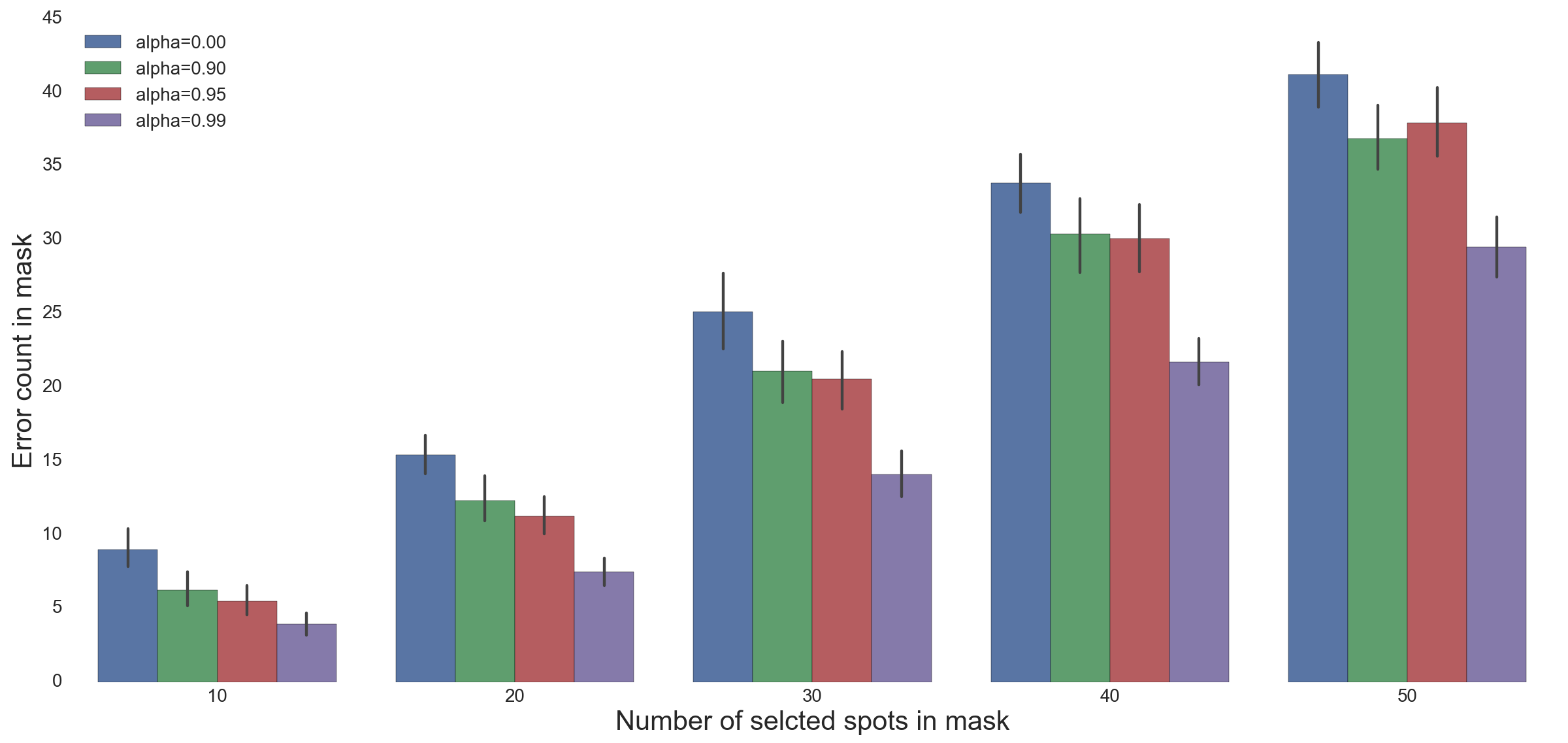}
	\caption{MNIST, Bilinear loss. We plot the number of erroneous test examples in the masked zone ($y$-axis) as a function of the zone's size ($x$-axis), mean over the $50$ repetitions per configuration with error bars showing the 95\% confidence interval.}
\label{fig-mnist-count}
\end{figure}

Results are shown in Fig. \ref{fig-mnist-count}. Clearly the number of test errors in the selected mask spots (the 'masked zone') is dramatically reduced when using our proposed loss function (\ref{eq:full-B}) with $\alpha >0$, as compared to the baseline models ($\alpha =0$). As expected, the total number of errors increases linearly with the size of the mask. At the same time the reduction in error count in the masked zone is attenuated by the value of the trade-off parameter $\alpha$, with higher values of $\alpha$ leading to fewer errors in this zone. Noticeably, as the number $n$ of selected spots in the mask is increased, a larger value of $\alpha$ is necessary in order to retain the reduction in the number of errors relative to the baseline. 

\begin{figure}[ht!]
    \centering
    \includegraphics[trim={0 0 0 0},clip, width=1.\textwidth]{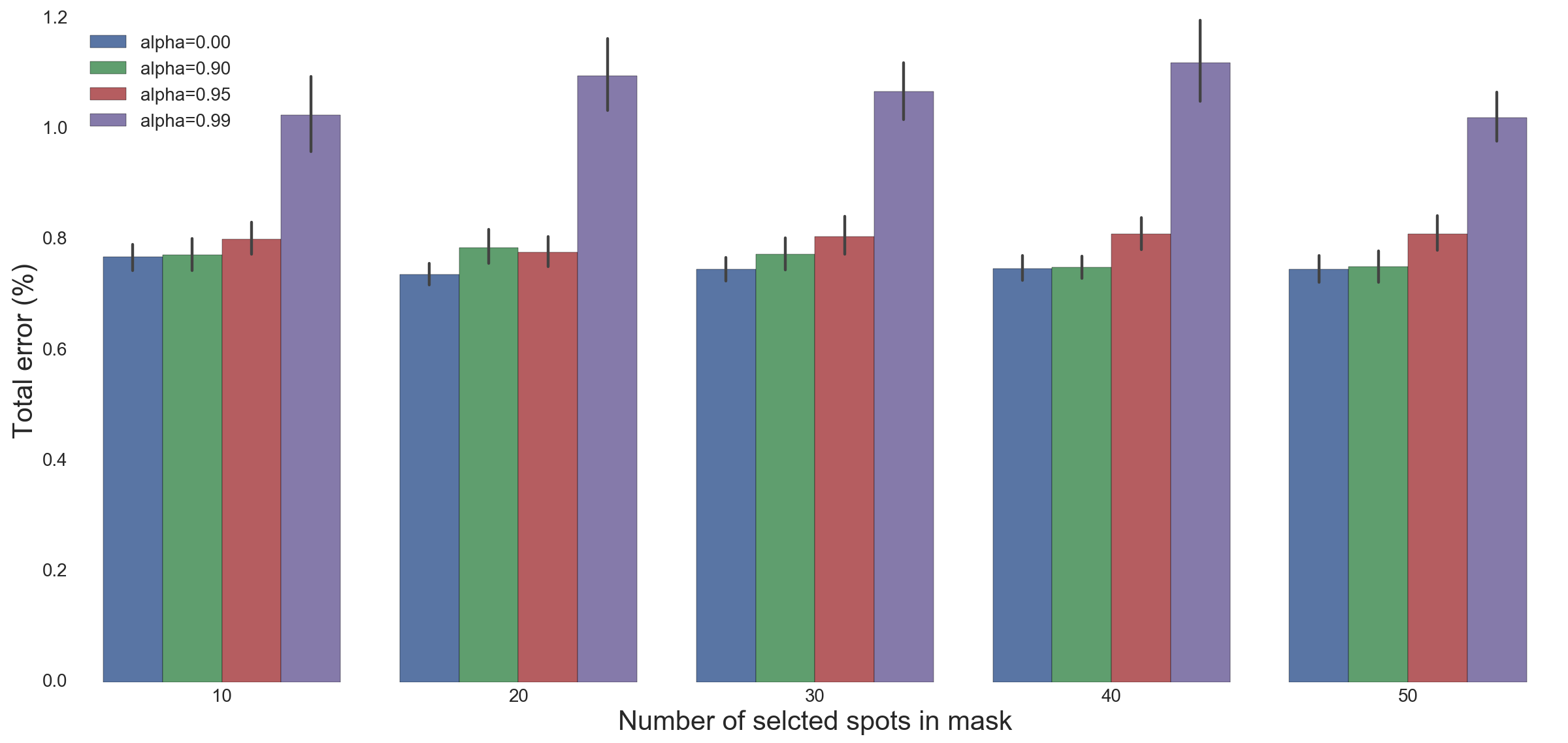}
	\caption{MNIST, Bilinear loss. We plot the percent of overall model error, mean over the $50$ repetitions per configuration with error bars showing the 95\% confidence interval.}
\label{fig-mnist-err}
\end{figure}

The overall model accuracy (Fig. \ref{fig-mnist-err}) changes very little for all but the largest value of $\alpha$. However, for $\alpha=.99$ we see a substantial increase in overall error. This value of $\alpha$ was also responsible for the most dramatic decrease in errors in the masked zone. Thus in this parameter value regime the network achieves an inferior overall solution, but succeeds in pushing most of the errors outside the masked zone. 

Overall, for intermediate values of the trade-off parameter $\alpha$, the number of errors in the masked zone is reduced dramatically, even for relatively large masks with as many as $40-50$ points, without an appreciable increase in overall error. This result demonstrates the feasibility of the proposed bilinear loss (when added to the regular cross-entropy loss), as a means of controlling the location of error without harming overall accuracy in deep learning models.

\subsubsection*{Log-Bilinear Loss.}

\begin{figure}[ht!]
    \centering
    \includegraphics[trim={0 0 0 0},clip, width=1.\textwidth]{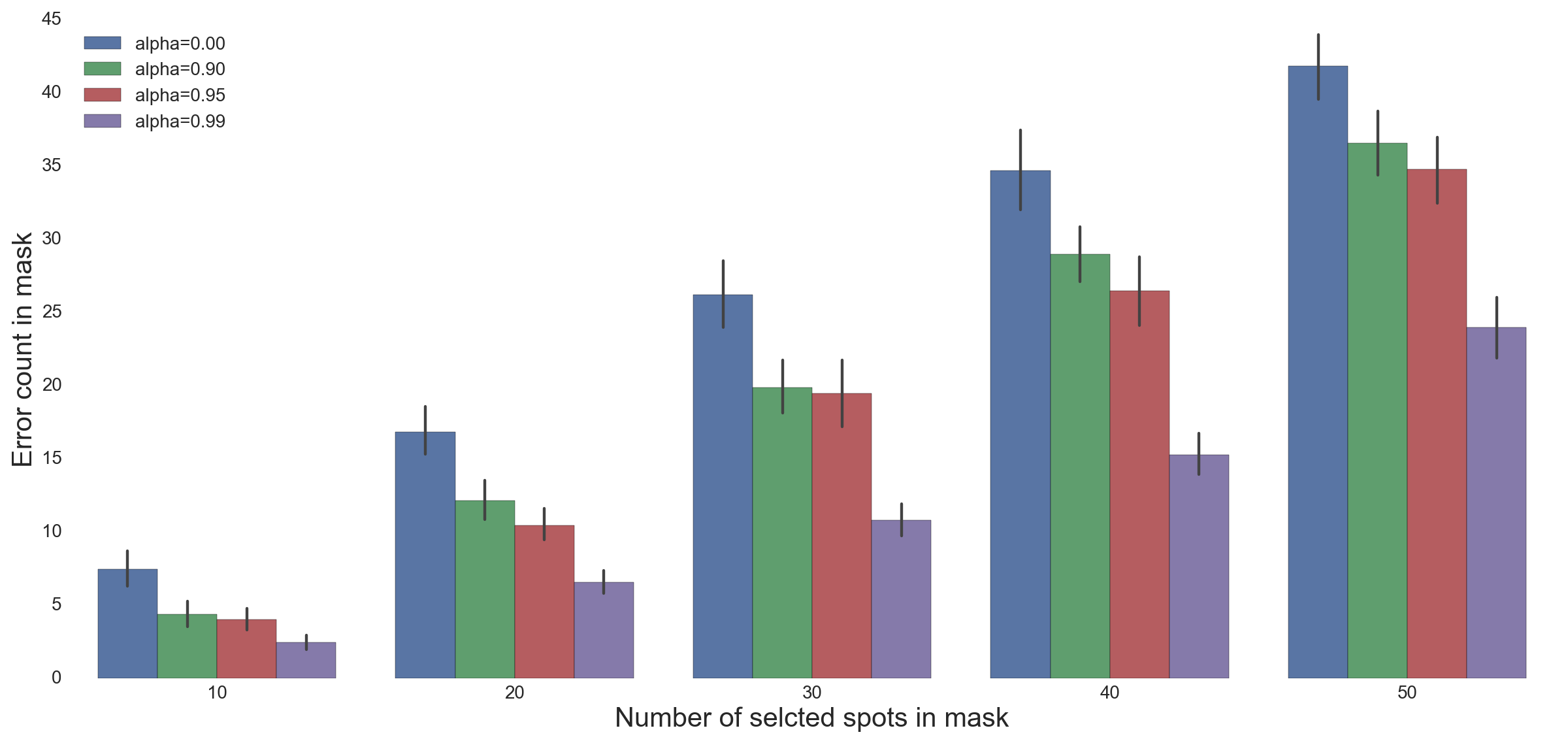}
	\caption{MNIST, Log Bilinear loss; see caption of Fig.~\ref{fig-mnist-count}.}
\label{fig-mnist-log-count}
\end{figure}

\begin{figure}[ht!]
    \centering
    \includegraphics[trim={0 0 0 0},clip, width=1.\textwidth]{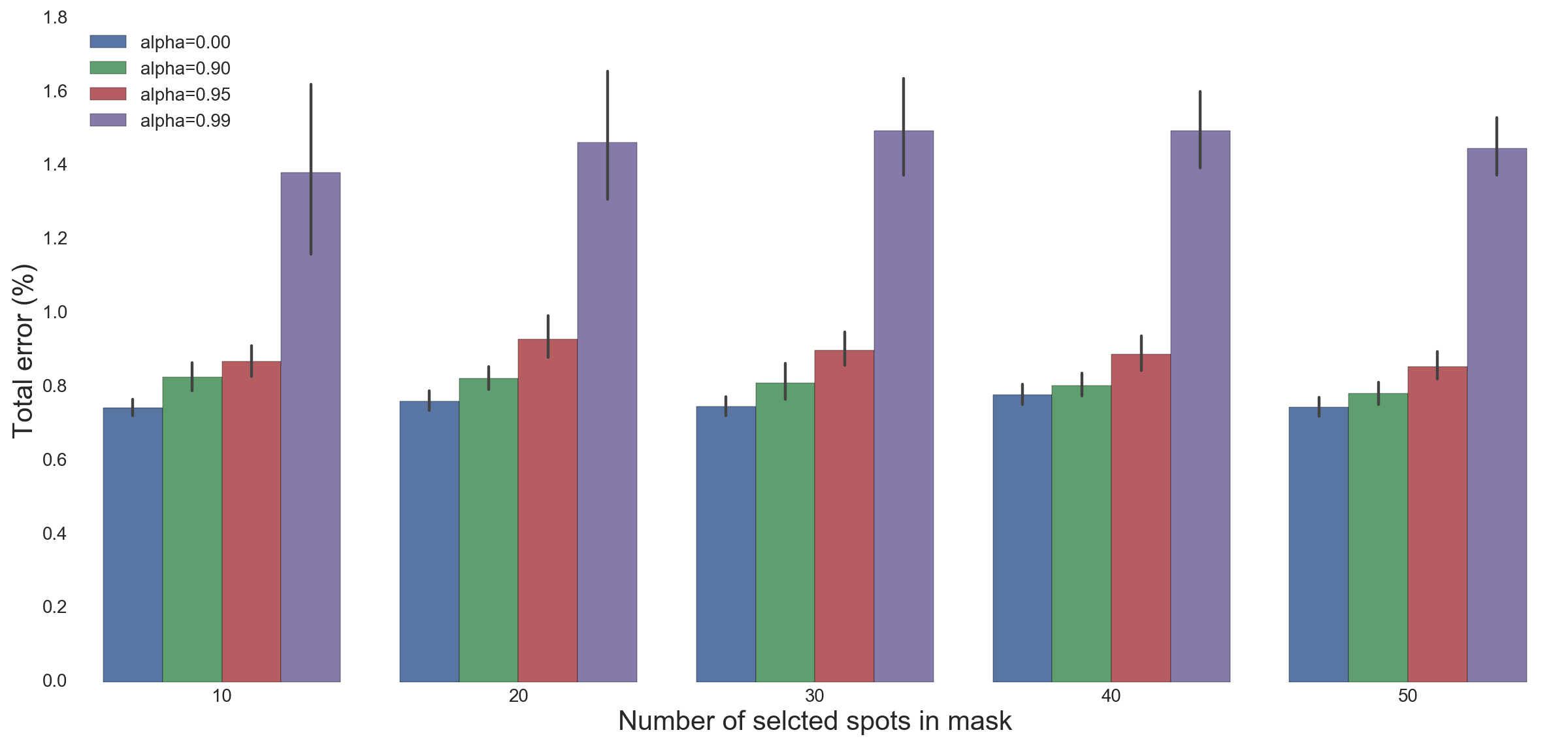}
	\caption{MNIST, Log Bilinear loss; see caption of Fig.~\ref{fig-mnist-err}.}
\label{fig-mnist-log-err}
\end{figure}

The results for the log-bilinear loss are qualitatively similar to the bilinear loss presented above. The reduction in the number of errors in the masked zone (Fig. \ref{fig-mnist-log-count}) relative to the baseline is larger than with the bilinear loss for small mask sizes ($n=10-20$), and comparable for the larger masks. The overall model accuracy (Fig. \ref{fig-mnist-log-err}) is, however, significantly worse than in the linear case, with small adverse effects showing already for small $\alpha$ values. 

Overall, the log-bilinear loss, much like the linear version, is able to reduce the number of errors in an MNIST model in a selected zone. However, the reduction in error in the masked zone when using the log-bilinear loss is more substantial, at a price of a worse model overall; this harmful effect is slight for small values of $\alpha$.

\subsection{CIFAR-10: results}

The CIFAR-10 dataset \cite{krizhevsky2009learning} is made up of $60,000$ small images ($32\times32$ color pixels), each belonging to one of $10$ classes (airplane, car, bird, cat, deer, dog, frog, horse, ship, truck). The data is divided into a training set of $50,000$ images, and a test set of the remaining $10,000$ images.

As in the MNIST case, the model we use (Table \ref{tbl-model-CIFAR}) is selected on the basis of typicality, accuracy, and relatively short training time to allow many models to be computed. Here, the model consists of three blocks of convolutional layers (each containing two convolutional layers, followed by max-pooling and dropout layers), and two fully-connected layers. Overall accuracy is approximately $92\%$ after $300$ epochs at most. ($300$ was set to be the maximal number of epochs; early stopping was employed when convergence was achieved earlier, which was often the case.)

\begin{table}[h!]
\centering
\caption{The model used in all CIFAR10/100 experiments}
\label{tbl-model-CIFAR}
\begin{tabular}{l|l|l}
  & Layer                 & \#params         \\ \hline \hline
1 & convolution (64; 3X3) & 1,792             \\
2 & convolution (64; 3X3) & 36,928            \\
3 & max-pooling(2X2)      &                  \\
4 & dropout               &                  \\ \hline

5 & convolution(128; 3X3)  & 73,856           \\
6 & convolution(128; 3X3)  & 147,584          \\
7 & max-pooling(2X2)      &                  \\
8 & dropout         &                        \\ \hline

9 & convolution(256; 3X3)  & 295,168          \\
10 & convolution(256; 3X3)  & 590,080         \\
11 & max-pooling(2X2)      &                 \\
12 & dropout         &                       \\ \hline

13 & fully-connected(1000)  & 257,000           \\
14 & dropout (CIFAR100 only) &                 \\
15 & fully-connected(1000)  & 1,001,000          \\
16 & dropout (CIFAR100 only) &                 \\
17 & fully-connected(10/100)   & 10,010/100,100  \\ \hline \hline 
  & total:                & 2,413,418 (CFIAR-10) \\ & & 2,503,508(CIFAR-100) 
\end{tabular}
\end{table}

\subsubsection*{Bilinear Loss.}

\begin{figure}[ht!]
    \centering
    \includegraphics[trim={0 0 0 0},clip, width=1.\textwidth]{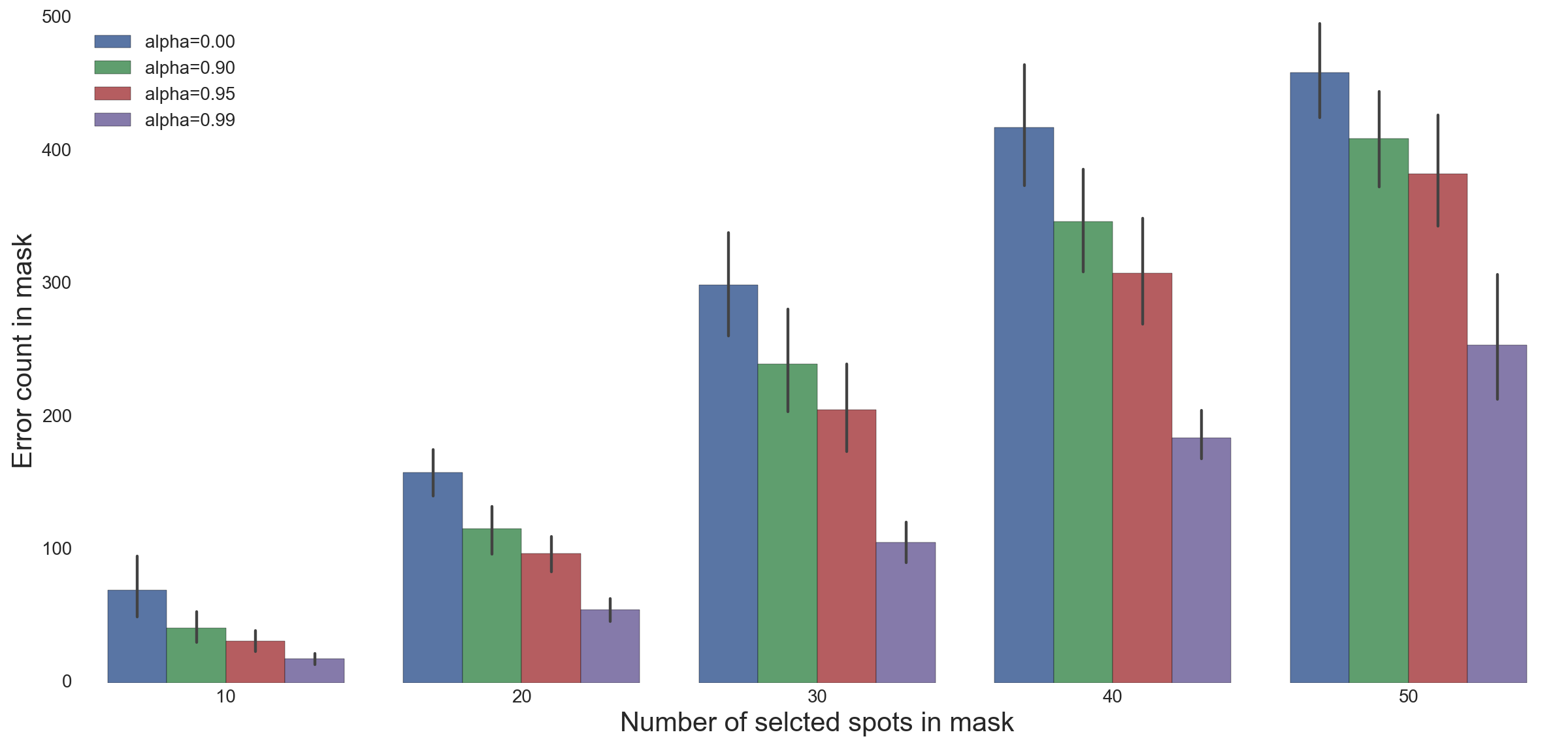}
	\caption{CIFAR-10, Bilinear loss. We plot the number of erroneous test examples in the masked zone ($y$-axis) as a function of the zone's size ($x$-axis), mean over the $10$ repetitions per configuration with error bars showing the 95\% confidence interval.}
\label{fig-cifar10-count}
\end{figure}

\begin{figure}[ht!]
    \centering
    \includegraphics[trim={0 0 0 0},clip, width=1.\textwidth]{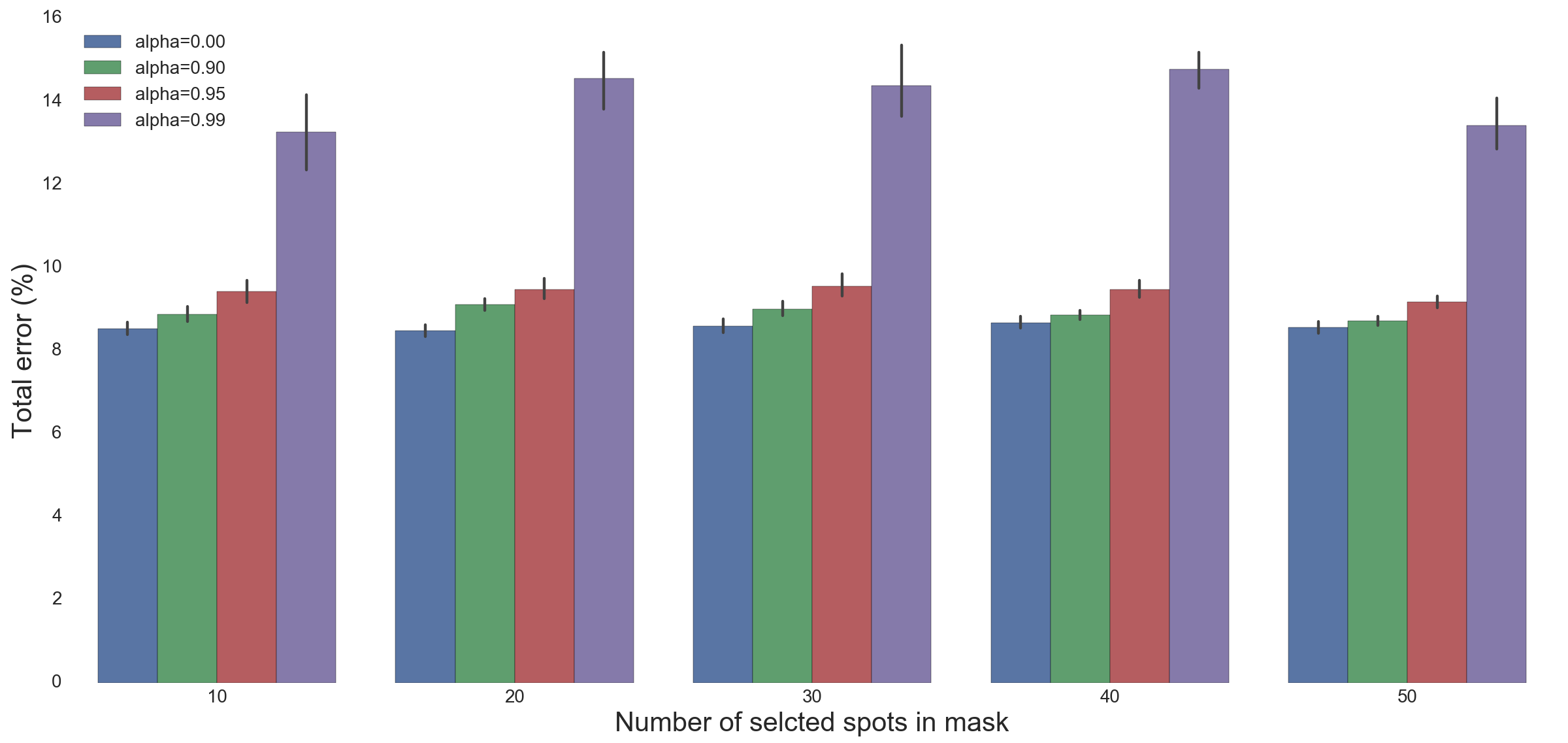}
	\caption{CIFAR-10, Bilinear loss. We plot the percent of overall model error, mean over the $10$ repetitions per configuration with error bars showing the 95\% confidence interval.}
\label{fig-cifar10-err}
\end{figure}
Results are shown in Fig. \ref{fig-cifar10-count}. Like before, the number of test errors in the masked zone is dramatically reduced when using our proposed loss function (\ref{eq:full-B}) with $\alpha >0$, as compared to the baseline models ($\alpha =0$). 

Compared to the MNIST results, for a small mask of $10$ locations, the reduction in error in the masked zone is appreciably larger for CIFAR-10. We attribute this to \textit{compulsory errors} -- errors which stem from true class overlap, and thus can't be overcome by changing the objective. Arguably, the MNIST dataset has a higher volume of these, explaining the better ability to control the error away from the mask when using the CIFAR-10 dataset. 

The overall model accuracy (Fig. \ref{fig-cifar10-err}) changes very little for all but the largest value of $\alpha$. Unlike the MNIST case, however, we see a real (albeit small) increase in overall error even for lower values of $\alpha$. For $\alpha=.99$ we again see a substantial increase in overall error. This value of $\alpha$ was also responsible for the most dramatic decrease in errors in the masked zone. Thus in this regime of parameter values the network achieves an inferior overall solution, but succeeds in pushing most of the errors outside the masked zone. 

Overall, for intermediate values of the trade-off parameter $\alpha$, the number of errors in the masked zone is reduced dramatically, even for relatively large masks with as many as $40-50$ points, without drastically harming the overall performance of the model.

\subsubsection*{Log-Bilinear Loss.}

\begin{figure}[ht!]
    \centering
    \includegraphics[trim={0 0 0 0},clip, width=1.\textwidth]{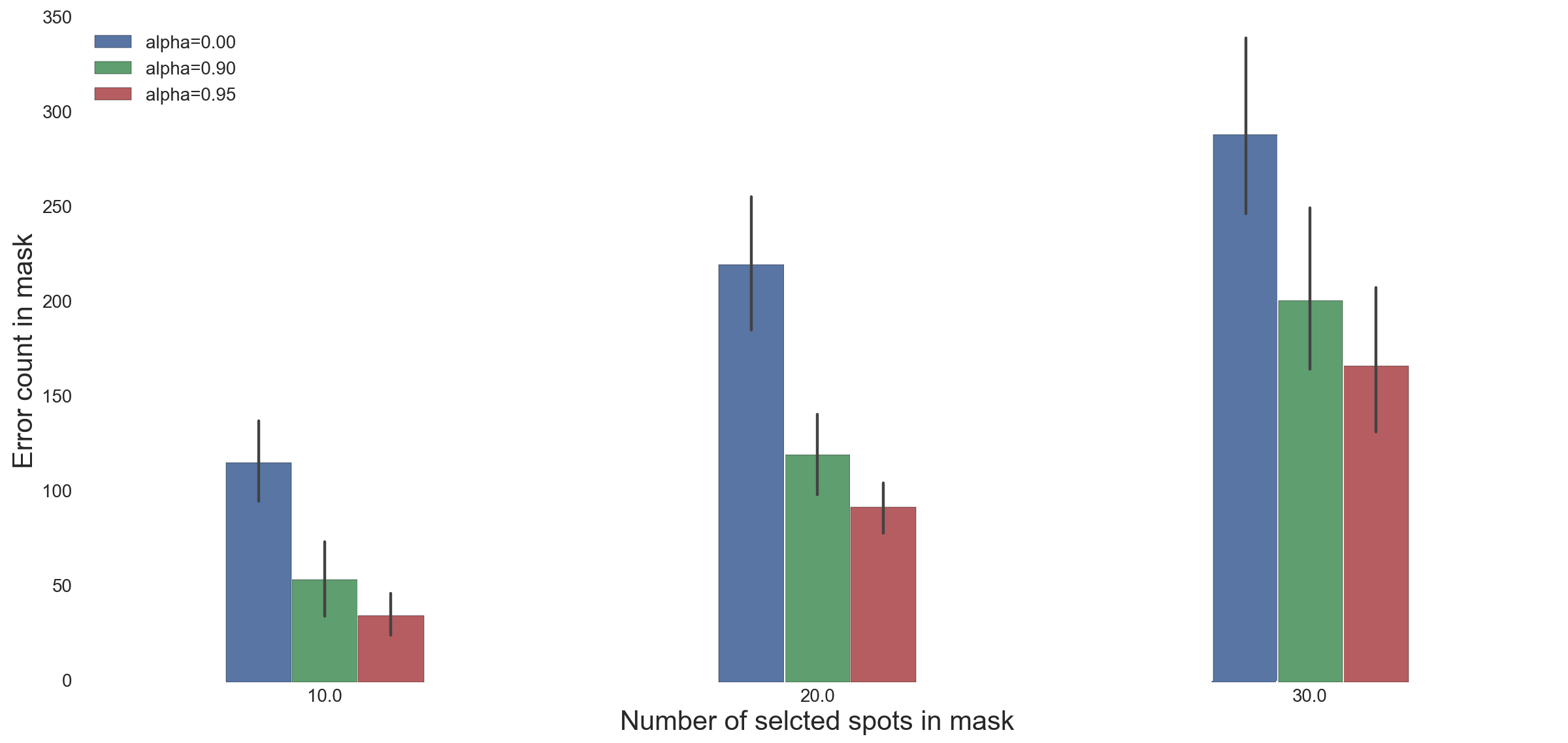}
	\caption{CIFAR-10, Log Bilinear loss; see caption of Fig.~\ref{fig-cifar10-count}.}
\label{fig-cifar10-log-count}
\end{figure}

\begin{figure}[ht!]
    \centering
    \includegraphics[trim={0 0 0 0},clip, width=1.\textwidth]{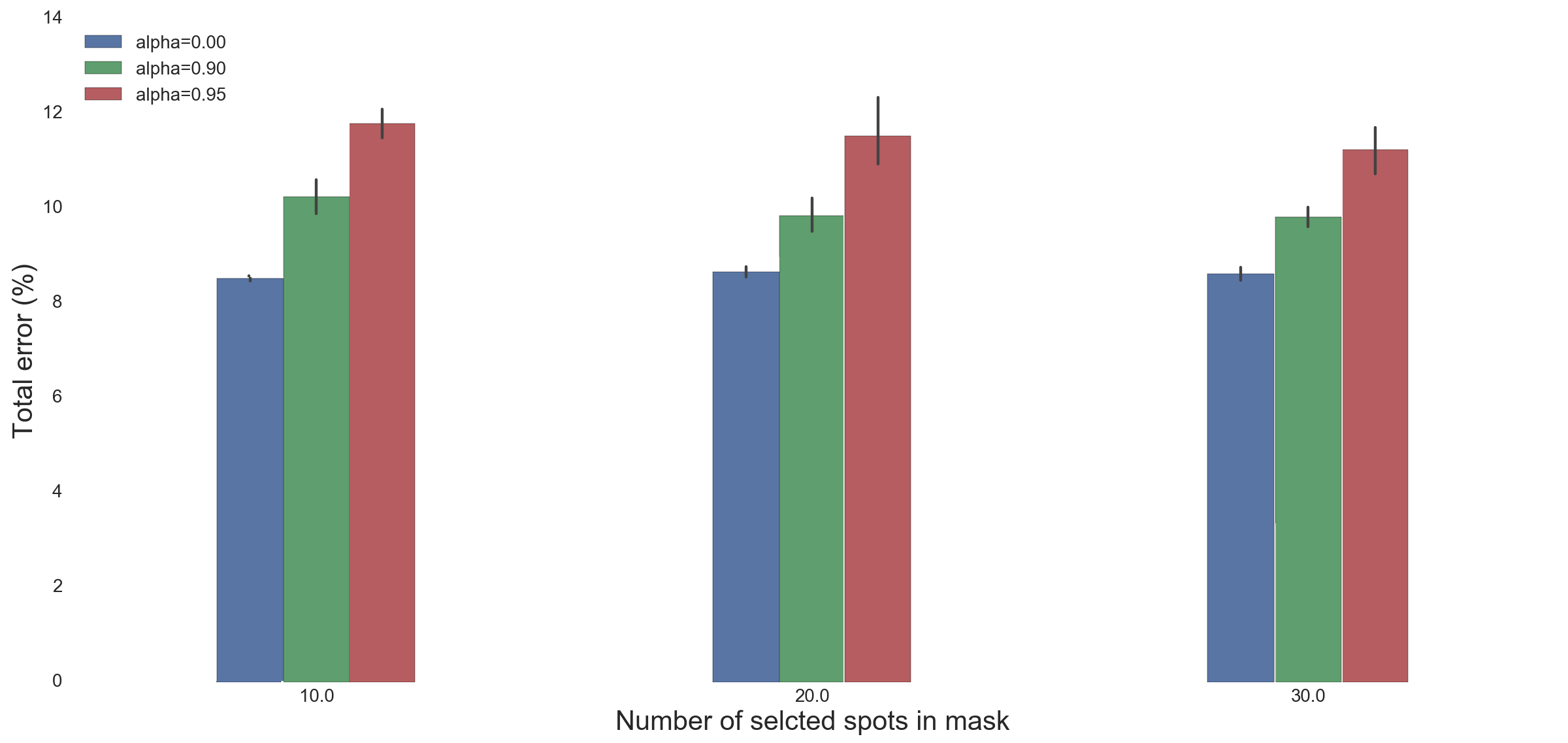}
	\caption{CIFAR-10, Log Bilinear loss; see caption of Fig.~\ref{fig-cifar10-err}.}
\label{fig-cifar10-log-err}
\end{figure}

The log-bilinear loss results show a large reduction in the error in the masked zone (Fig. \ref{fig-cifar10-log-count}), but at the same time an increase in overall model error (Fig. \ref{fig-cifar10-log-err}) already for small masks and smaller values of $\alpha$. 

We show data in this case (Fig. \ref{fig-cifar10-log-count} and \ref{fig-cifar10-log-err}) for masks of size up to 30, and $\alpha$ values of up to $0.95$. The log-bilinear loss is less effective than the bilinear loss already in this regime, and is not of practical use for this dataset with larger masks. It would seem that for this dataset the bilinear loss produces better results overall in terms of control of error on the one hand, while maintaining reasonable overall model accuracy on the other.

\section{Hierarchical classification: empirical evaluation}
\label{sec:hierar}

When the data is organized hierarchically in a tree, a natural order in imposed on the errors that the classifier can make. In some domains it may be beneficial to penalize for errors based on the distance between the node of the true label and the node corresponding to the erroneous label, where the distance reflects the lowest node in the tree that is ancestral to both nodes, since the detrimental effect of errors may be proportional to this distance.

In this section we investigate this scenario using the hierarchical CIFAR100 dataset \cite{krizhevsky2009learning}; this dataset contains $100$ classes, which are grouped into $20$ coarse-grained super-classes (for example, the \textit{reptile} super-class contains: crocodile, dinosaur, lizard, snake, and turtle; the \textit{insects} super-class contains: bee, beetle, butterfly, caterpillar, and cockroach). %In line with the idea that the loss function should include a larger penalty when two labels are less similar to each other, thIn the following experiments we use the bilinear (and log bilinear) loss in order to contain the error in the correct super-class.

Hierarchical structure is often utilized by classifiers in order to improve the regular notion of classification \cite{tsochantaridis2005large,gopal2013recursive}. The objective here is to generate a classifier that, when wrong, will be more likely to err by choosing another class from the same super-class containing the true label.

We achieve this goal by using the bilinear and log-bilinear loss functions defined in (\ref{eq:full-B}) and (\ref{eq:full-LB}), where the elements of the penalty matrix $A=\{ a_{i,j}\}$ are defined as follows:
\begin{equation*}
	a_{i,j} = \begin{cases} 
    	0 & i = j \\
        1 & \text{i and j are in the same super-class} \\ 
        5 & \text{o.w.}
    \end{cases}
\end{equation*}
This penalty matrix reflects the notion that we would much rather err within the correct super-class. 

As in the above experiments, we test multiple  values of the trade-off parameter $\alpha$ for each of the bilinear and log-bilinear models, and $10$ repetitions of each combination. The results show that when using either the bilinear (Table \ref{tbl-c100-linear}) or the log-bilinear (Table \ref{tbl-c100-log}) loss functions, we are able to reduce the error for coarse-grained classification (into super-classes) without hindering the overall model accuracy.

\begin{table}[h!]
\centering
\caption{CIFAR-100, Bilinear loss. $\alpha$: the trade-off parameter (see (\ref{eq:full-B})). Total error: the mean overall test set error. Total error coarse: the mean overall test set error when using the coarse classification into the $20$ super-classes. Relative error with correct super-class label: the mean percent of errors which are assigned a label in the correct super-class. } 
\label{tbl-c100-linear}
\begin{tabular}{P{2.5cm}||P{2.5cm}|P{3.5cm}||P{2.5cm}}
$\alpha$ & total error & total error coarse & relative error with correct super-class label  \\ \hline\hline
0     & 37.36       & 25.45   & 30.88                                                   \\ \hline
0.05  & 37.41        & 25.33   & 31.22                                                      \\ \hline
0.10   & 37.33       & 25.27   & 31.63                                                       \\ \hline
0.25  & 37.13       & 24.62  & 32.92                                                       \\ \hline
0.50   & 36.93        & 24.01      & 34.61                                                   \\ \hline
0.75  & 38.33       & 24.52     & 35.76                                                     \\ \hline
0.90   & 41.41        & 26.45   & 35.98                                                      \\ \hline
0.95  & 44.57       & 28.82  &  35.27                                              
\end{tabular}
\end{table}

\begin{table}[h!]
\centering
\caption{CIFAR-100, Log-Bilinear loss. See caption in Table \ref{tbl-c100-linear}.}
\label{tbl-c100-log}
\begin{tabular}{P{2.5cm}||P{2.5cm}|P{3.5cm}||P{2.5cm}}
$\alpha$ & total error & total error coarse & relative error with correct super-class label  \\ \hline\hline
0     & 37.30 & 25.43 & 30.88   \\ \hline
0.05  & 37.23 & 25.12 & 31.55   \\ \hline
0.10  & 37.25 & 24.92 & 32.20   \\ \hline
0.25  & 37.61 & 24.71 & 33.60   \\ \hline
0.50  & 38.52 & 24.52 & 36.58   \\ \hline
0.75  & 41.53 & 25.73 & 38.32   \\ \hline
0.90  & 46.18 & 28.26 & 38.87   \\ \hline 
0.95  & 48.98 & 30.31 & 38.21                                    
\end{tabular}
\end{table}

Specifically, the coarse-grained overall error is reduced from $25.45\%$ to $24.01\%$ when using the bilinear loss with $\alpha = 0.5$, and to $24.52\%$ when using the log-bilinear loss with the same values of $\alpha$. Interestingly, in the latter case the overall model accuracy is slightly (but significantly) reduced as well. 

In the baseline model ($\alpha=0$), $30.88\%$ of all mislabeled examples are mislabeled into the correct super-class (i.e. assigned one of the other $4$ classes that form the same super-class as the correct label). \footnote{This value, rather than the expected $\frac{4}{99}$ reflects the hierarchical structure that exists in this dataset, as captured in the super-class labels.}  
%The percent of erroneous classification labels which are assigned to the correct super-class is on average $30.88\%$ for the baseline model (recall that $\alpha=0$ reduces to the regular cross-entropy loss).
When using the bilinear loss, the percent of erroneous classification labels that fall within the correct super-class in increased to $34.61\%$ with $\alpha=0.5$. Likewise, the value increases to $36.58\%$ with the log-bilinear loss. However, in the latter case the total error both in the regular ($100$ classes) and coarse classification ($20$ super-classes) is somewhat higher.   

It seems that for this task the bilinear loss is more effective than the log-bilinear loss. The reason for this may be that the many misclassification errors are evaluated with low confidence, which gives a relatively flat response for the $100$ classes. Thus, while the bilinear loss accumulates these errors, the log-bilinear loss adds up values which are very small (the sum of $log(1-\epsilon)$). 

We note that while in the MNIST/CIFAR10 experiments we generate masks to direct the error away from as many as $50$ out of the $90$ possible error types, here only $400$ out of the possible $9,900$ possible errors are within super-class errors (each super-class is of size $5$). Remarkably, even in this extreme case we see a clear effect of the error being controlled as defined in the bilinear loss.  

\subsubsection*{Small sample}

In the final experiment presented here, we test the effect of the bilinear loss in the small-sample case. From each of the $20$ super-classes of the CIFAR-100 dataset, we select a class based on the highest super-class-typicality, which we define as the percent of the errors for each class that are assigned to another class within the same super-class, in the baseline model. 

Each of the $20$ selected classes is down-sampled to $0, 10,$ or $50$ training examples. In the $10$ and $50$ cases, the small sample is then duplicated back to the original sample size. Models are then trained using the bilinear loss, using the matrix described above (penalizing differentially errors in the correct and wrong super-class), with values of the trade-off parameter $\alpha$ in $\{0, 0.25, 0.5\}$. 

The results (Table \ref{tbl-c100-small-sample}) show a small improvement in model accuracy - the accuracy of the small-sample classes, and small improvement in the assignment of the small-sample classes to the correct super-class for samples of size $N=10, 50$.  Interestingly, even when the selected classes are not seen at at all during training ($N=0$), and even for the baseline model ($\alpha=0$), almost half ($49.42\%$) of the test examples from the selected classes are assigned to the correct super-class. While somewhat expected, since the most typical classes are used, this again mostly reflects the visually informative super-class structure in the CIFAR-100 dataset.

\begin{table}[h!]
\centering
\caption{CIFAR-100 with selected small-sample classes. N: the number of samples in each of the $20$ small-sample classes. alpha: the value of the trade-off parameter $\alpha$. model accuracy: the overall accuracy of the model. small sample accuracy: the mean accuracy for the $20$ small sample classes. small sample super-class accuracy: the percent of test examples from the small-sample classes that are assigned a class in the correct super-class} 
\label{tbl-c100-small-sample}
\begin{tabular}{P{1.0cm}|P{1.5cm}|P{2.5cm}|P{2.5cm}|P{2.5cm}}
N  & $\alpha$    &  model accuracy &  small sample accuracy &  small sample super-class accuracy \\ \hline\hline 
0 & 0 &              54.92 &             0.00 &            49.42 \\
  & 0.25 &              54.94 &             0.00 &            49.24 \\
  & 0.50 &              54.91 &             0.00 &            50.54 \\ \hline
10 & 0 &             51.76 &             4.23 &            47.82 \\
  & 0.25 &              52.33 &             4.45 &            49.20 \\
  & 0.50 &              52.72 &             6.02 &            50.14 \\ \hline 
50 & 0 &             56.88 &            21.98 &            56.43 \\
  & 0.25 &              56.87 &            22.43 &            56.85 \\
  & 0.50 &              57.11 &            23.47 &            58.52 \\
\end{tabular}
\end{table}

\comment{
\section{Future Work}
\subsection{Learning with Hierarchies}
A particularly intriguing application of the Log Bilinear loss is for forcing hierarchies in the layers of a deep model. It is often the case (especially in real-world datasets), that objects being classified have labels with an hierarchical structure. For instance, a car of a specific model is also just a car, a vehicle, and a man-made object. 

The idea behind learning with hierarchies and the bilinear loss is to require that specific layers represent objects in a way that allows a readout for a certain level in the hierarchy. This is done by setting the specific loss for predicting an erroneous class within the given level of the hierarchy to a lower value than any other erroneous class. This idea is clearly represented by a block-diagonal confusion matrix in the bilinear loss. The final objective function is then the sum of the confusion matrix losses, applied at the various layers of the deep model. 

More concretely, we define a more general form of the Bilinear Loss (\ref{eq:full-B}), as follows:

\begin{equation}
	L_{CE+B} = (1-\alpha) L_{CE} + \alpha y^T A(\theta_1, ...,\theta_k) \hat{y} 
\end{equation}

\noindent where $A(\theta_1, ...,\theta_k)$  in the $i,j$'th position contains a \textbf{function} of the correct and assigned classes, as before, but also a set of extra arbitrary parameters:

\[ a(i, j) = f(i,j, \theta_1, ...,\theta_k) \]

For simplicity, we now suppose that the training set is labeled with class labels at only two different degrees of granularity, $y_1$ and $y_2$ and that $y_1$ is an indicator vector (fine-scale), and $y_2$ encodes the course-labels as a uniform distribution over subsets of the the fine-scale classes (using the car example above -- a car is encoded as a uniform distribution over car models). 

We construct a deep model with $k$ output levels (i.e. $k$ predictors $\hat{y}_1, ... , \hat{y}_k$). Then the total loss is defined as:

\[ 
	\frac{1}{k} \sum_{l=1}^{k} (1-\alpha) L_{CE} + \alpha y^T A(l) \hat{y_l} 
\]

\noindent where:

\[ 
	A(l)[i, j] = \begin{cases} 
    	1 & \text{i and j not in same course-grained class} \\
        \frac{k}{l} & o.w.  
    \end{cases}
\]

\noindent meaning that the penalty for inside-course-classification errors increases linearly throughly the $k$ levels of the output, reaching eventually the same status as any other error, in the final output layer only. 

At the test stage, the model can then be used to probe for the class of a test-object, and depending on the required level in the hierarchy, processing can stop at the right depth in the model, and use the correct readout (meaning effectively that courser classification can be preformed faster). 
}

\section{Conclusions}

Often in real-world classification tasks, the cost of an error depends on the identity of the correct label as well as that of the misclassification target. The cross-entropy objective most commonly used in classification with deep learning models does not accommodate this in a natural way.

In this paper we present the bilinear and log-bilinear loss functions, which directly add to the regular cross-entropy loss a component that depends on the true and wrongly assigned labels. We evaluate these formulations extensively using standard models and benchmark datasets. 

First, we show that with the MNIST/CIFAR-10 datasets we are able to direct error away from a randomly chosen mask of up to $50$ out of the $90$ possible errors (the non-diagonal elements of the $10$ by $10$ confusion matrix), while preserving the overall model accuracy.

Next, we show that in the hierarchically annotated CIFAR-100 dataset, these methods are able to help contain error to within the correct super-class, thus producing models with the same overall accuracy as the baseline model, but with a higher inclination to choose a label from sibling classes (as defined by the super-class) when mistakes occur. 

Future work will focus on extending this "hierarchical learning" by coercing the layers of a deep model to gradually focus on higher and higher levels of a fuzzy hierarchy.  

\bibliographystyle{splncs03}
\bibliography{lib}

\end{document}